# Graphs Unveiled: Graph Neural Networks and Graph Generation

LÁSZLÓ KOVÁCS
University of Miskolc, Hungary
Institute of Information Technology
kovacs@iit.uni-miskolc.hu

ALI JLIDI
University of Miskolc, Hungary
Institute of Information Technology
jlidi.ali@student.uni-miskolc.hu

**Abstract:** One of the hot topics in machine learning is the field of GNN. The complexity of graph data has imposed significant challenges on existing machine learning algorithms. Recently, many studies on extending deep learning approaches for graph data have emerged. This paper represents a survey, providing a comprehensive overview of Graph Neural Networks (GNNs). We discuss the applications of graph neural networks across various domains. Finally, we present an advanced field in GNNs: graph generation.

## 1. Introduction

Embarking on the exploration of machine learning applied to graphs [1] invites us into a realm where graphs, representing connections between objects (nodes), become a universal language for deciphering complex systems [2]. For instance, in a social network graph, individuals are nodes, and friendships are edges. The power of this concept becomes evident in historical studies, like Wayne W. Zachary's analysis of a karate club's dynamics [3], predicting factional splits based on the graph structure. What makes graphs versatile is their ability to represent various interactions, be it in social networks, biology, or even telecommunications.

Now, as we step into the world of machine learning, graphs become more than visual representations. They serve as a mathematical foundation enabling us to analyze and understand intricate relationships within real-world complex systems. With the surge in available graph-structured data from sources like social networks, scientific initiatives, and interconnected devices, the challenge lies in unleashing the potential of this data.

At the same time, the second part of our exploration looks at why people are paying a lot of attention to studying graphs with machine learning. Graphs, which are like maps showing connections between things, are powerful tools. They help us understand many different things like how people interact in social networks, how proteins work together, or how information is organized in knowledge graphs. The advent of Graph Neural Networks (GNNs), rooted in the history of neural networks for graphs and inspired by the success of convolutional neural networks (CNNs), has revolutionized graph analysis. GNNs collectively aggregate information from graph structures, enabling tasks like node classification [4] and link prediction [5].

This paper unfolds the motivations behind GNNs [6]. It explores the role of graph representation learning in enhancing traditional machine learning approaches. The review provides a comprehensive understanding of GNNs, categorizes them into different groups, and delves into their applications across various domains. It also identifies open problems for future research, making this exploration a roadmap for understanding and advancing machine learning on graphs.

## 2. Background survey

### 2.1. Graph description

Before we dive into talking about machine learning on graphs, let's first explain what we mean by "graph data" in simpler terms. A graph, formally represented as G = (V, E) as it shown in figure 1, is made up of nodes (V) and edges (E) connecting these nodes. An edge from one node (u) to another (v) is written as (u, v) ∈ E. Usually, we're dealing with straightforward graphs, where there's at most one connection between each pair of nodes, no self-connections, and all connections are two-way, meaning (u, v) ∈ E is the same as (v, u) ∈ E.

We often use an adjacency matrix (let's call it A) to represent graphs [7]. This matrix, with dimensions |V|×|V|, helps us see which nodes are connected. If there's a connection between nodes u and v, A [u, v] equals 1; otherwise, it's 0. If the graph only has undirected connections, the matrix is symmetrical, but if the graph is directed connections, it might not be. Some graphs also have "weighted" edges, meaning the connections aren't just 0 or 1, but can be any real number. For instance, in a graph showing how proteins interact, a weighted edge could tell us how strong the connection is between two proteins.

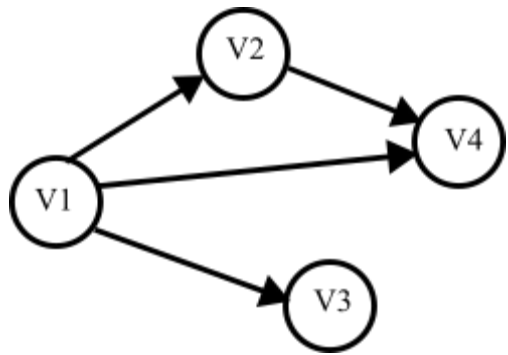

*Figure 1: Graph G = (V, E)*

#### 2.1.1 Graph type and scale

Graphs with complex types could provide more information on nodes and their connections. Graphs are usually categorized as:

- **Directed/Undirected Graphs.** Edges in directed graphs are all directed from one node to another, which provides more information than undirected graphs. Each edge in undirected graphs can also be regarded as two directed edges.
- **Homogeneous/Heterogeneous Graphs.** Nodes and edges in homogeneous graphs have same types, while nodes and edges have different types in heterogeneous graphs. Types for nodes and edges play important roles in heterogeneous graphs and should be further considered.
- **Static/Dynamic Graphs.** When input features or the topology of the graph vary with time, the graph is regarded as a dynamic graph. The time information should be carefully considered in dynamic graphs.

These categories operate independently, allowing for combinations; for instance, a dynamic directed heterogeneous graph is one possible combination. Various other graph types, like hypergraphs and signed graphs designed for specific tasks, exist. Although not exhaustively listed here, the primary consideration lies in acknowledging the additional information offered by these graphs.

Concerning graph scale, the determination of what qualifies as "small" or "large" lacks a fixed criterion. The definition evolved with advancements in computational devices, such as the speed and memory of GPUs. In this paper, a graph earns the classification of large-scale when the device encounters challenges in storing and processing its adjacency matrix or graph Laplacian, which has a space complexity of $O^{(n^2)}$. In such instances, methodologies like sampling may become necessary.

$*O^{(n^2)}$.: n represents the number of nodes in the graph, the space complexity indicating that the space required grows quadratically with the number of nodes in the graph

### 2.2. Application area

Exploring the application areas of graph-based machine learning unveils a spectrum of diverse domains where these techniques prove instrumental. Social network analysis [8] stands out as a primary application, employing graphs to represent relationships among individuals. This enables the discernment of communities, influential nodes, and patterns within social structures. In recommendation systems, graphs model user-item interactions, facilitating the provision of personalized content recommendations based on shared preferences. Within the realm of biology, graph-based models contribute to understanding molecular interactions, where nodes signify biological entities and edges convey relationships. Transportation and logistics benefit from graph algorithms, optimizing routes and enhancing network connectivity. Notably, in graph generation, these techniques play a pivotal role in creating realistic graph structures, such as generating realistic social networks or molecular structures, adding an innovative dimension to the field by creating synthetic yet representative data for various applications. This multifaceted applicability underscores the adaptability of graph-based machine learning, offering inventive solutions across a spectrum of disciplines.

**Table 1 :** Application of graph neural networks

| Area | Application |
| --- | --- |
| Graph Mining | Graph Matching |
| | Graph Clustering |
| Physics | Physical Systems Modeling |
| Chemistry | Molecular Fingerprints |
| | Chemical Reaction Prediction |
| Biology | Protein Interface Prediction |
| | Side Effects Prediction |
| | Disease Classification |
| Knowledge Graph | Knowledge Base completion |
| | Knowledge Graph alignment |
| Generation | Graph Generation |
| Combinatorial Optimization | Combinatorial Optimization |
| Traffic Network | Traffic State Prediction |
| Recommendation Systems | User-item Interaction Prediction |
| | Social Recommendation |
| Others | Stock Market |
| | Software Defined Networks |
| | AMR Graph to Text |
| Image | Social Relationship Understanding |
| | Image Classification |
| | Visual Question Answering |
| | Object Detection |
| | Interaction Detection |
| | Region Classification |
| | Semantic Segmentation |
| Text | Text Classification |
| | Sequence Labeling |

| |
|---|
| Neural Machine Translation |
| Relation Etraction |
| Event Extraction |
| Fact Verification |
| Question Answering |
| Relational Reasoning |

### 2.3. Dynamic operations

In the exploration of dynamic operations [9], our attention turns to the temporal dimension of graphs, investigating how they evolve over time. Dynamic operations refer to the study and manipulation of graphs with changing structures, capturing variations in connections, entities, or attributes across different time points. This temporal layer adds complexity to traditional static graphs, offering insight into the dynamic nature of real-world systems, where relationships between entities can undergo transformations over time. Whether tracking shifts in social network interactions, observing changes in biological processes, or analyzing fluctuations in transportation networks, this section delves into the methodologies and considerations associated with dynamic graph operations. It aims to elucidate the mechanisms employed to navigate and comprehend the temporal evolution of graph structures within the context of complex systems.

A dynamic graph model is defined as $G_t = (V_t, E_t)$, where it outlines the state of the graph (comprising nodes and edges) at a specific time moment [10], denoted as t. Both directed and undirected dynamic graphs find representation in various existing models, be they discrete or continuous. In discrete models, periodic snapshots are taken at fixed intervals, such as every 30 minutes, day, or week. This approach offers precise mappings at specific time points and allows approximations for states at other instances based on factors like time or changes. Conversely, continuous models meticulously track every change, presenting an accurate graph state for any given instant. The subsequent sections categorize the pertinent literature into four distinct types, providing a detailed description of each.

## 3. General design pipeline of GNNs

### 3.1. Graph learnings tasks:

Graph learning tasks involve three main types: node-level tasks focus on individual nodes, edge-level tasks examine connections between nodes, and graph-level tasks address the properties of the overall graph.

- **Node-level tasks** in graph learning, such as classification, regression, and clustering, focus on categorizing nodes, predicting continuous values, and grouping similar nodes, respectively.
- **Edge-level** tasks involve classifying edge types and predicting the existence of edges between specified nodes.
- **Graph-level tasks** encompass graph classification, graph regression, and graph matching, requiring the model to learn representations for entire graphs.

From a supervision standpoint, graph learning tasks can be categorized into three training settings:

- **Supervised setting** utilizes labeled data for training.
- **Semi-supervised setting** involves a mix of labeled and unlabeled nodes, often seen in tasks like node and edge classification.
- **Unsupervised setting** relies solely on unlabeled data, suitable for tasks like node clustering.

### 3.2. Basic Design concept of GNN's

In graph-based learning, extracting meaningful representations as shown in figure 2 involves deriving a node feature vector and adjacency matrix. The node feature vector captures essential characteristics of individual nodes, serving as a condensed representation [11]. Simultaneously, the adjacency matrix encapsulates the relationships between nodes, reflecting the graph's connectivity structure. With these components in hand, embedding techniques can be applied to map nodes into a continuous vector space, facilitating efficient representation learning. Message passing algorithms [12], leveraging the adjacency matrix, then enable nodes to exchange information and refine their embeddings through iterative communication within the graph structure.

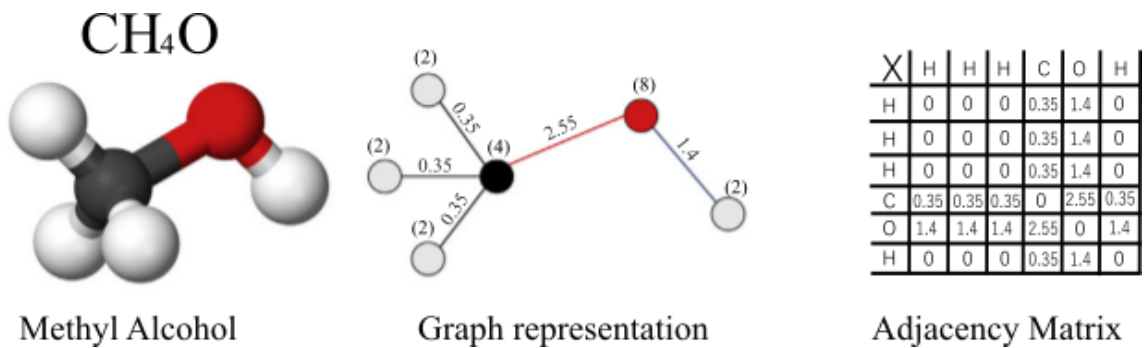

**Figure 2:** Extracting adjacency matrix from a graph

#### 3.2.1　　　Adjacency Matrix(A):

The adjacency matrix represents the connections between nodes in the graph. For a graph with N nodes, the adjacency matrix A is a matrix, where $A_{ij} = 1$ if there is an edge between node (i) and node (j), and =0 otherwise. As shown in figure 3

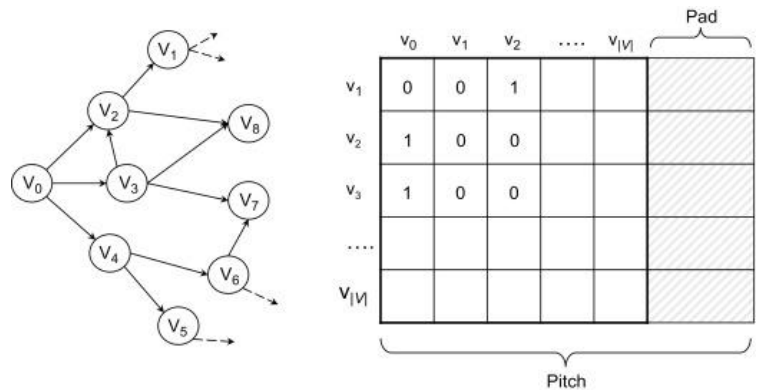

**Figure 3:** mapping an adjacency matrix.

#### 3.2.2　　　Node Embeddings (H):

Node embedding is a pivotal concept in graph representation learning, aiming to transform individual nodes into continuous vector representations. By capturing the inherent structural and relational information within a graph using different computational modules as shown in figure 4, node embeddings facilitate the translation of complex network structures into a more interpretable and computationally efficient format. These embeddings serve as compact yet informative representations, enabling downstream tasks such as node classification, clustering, and link prediction in a variety of applications, from social networks to biological systems.

$$\text{New Embeddings} = \text{adj\_matrix} \times X$$

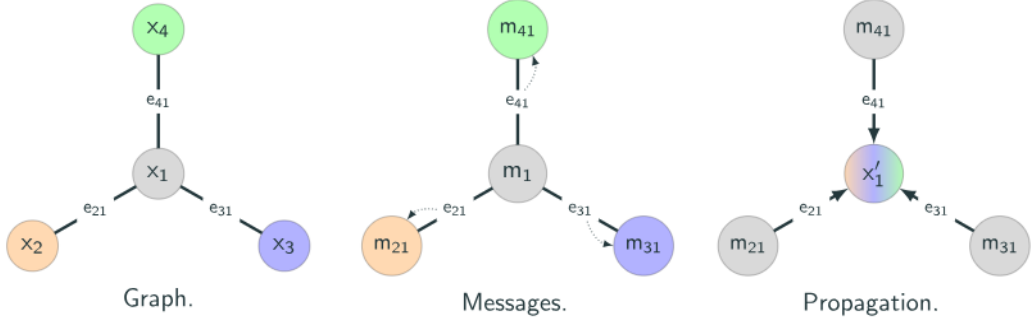

**Figure 4:** node embedding using message passing.

*3.2.3    Message passing algorithm:*

The fundamental graph neural network (GNN) model can be explained in various ways. Researchers have derived this core GNN model as an extension of convolutions to non-Euclidean data [Bruna et al., 2014], a differentiable version of belief propagation [Dai et al., 2016], and through analogies to traditional graph isomorphism tests [Hamilton et al., 2017b]. Regardless of the specific motivation, a defining characteristic of GNNs is their use of neural message passing. In this approach, vector messages are exchanged between nodes and updated using neural networks [Gilmer et al., 2017].

In this paper, we delve into the foundational aspects of this neural message passing framework. Our focus will be on the message passing mechanism itself, while detailed discussions about training and optimizing GNN models are reserved for later. This chapter primarily explores how an input graph $G = (V, E)$, along with a set of node features $X \in R_d \times |V|$, can be utilized to generate node embeddings $zu$ for all nodes $u \in V$. Moreover, we will also explore how the GNN framework can be employed to generate embeddings for subgraphs and entire graphs.

\* $X \in R_d \times |V|$,: Denotes a real-valued matrix of dimensions $d \times |V|$. where "d" is the number of features associated with each node and |V| is the number of nodes in the graph.

During each iteration of message-passing within a Graph Neural Network (GNN), the hidden embedding $h_u^{(k)}$[13] corresponding to every node u in the set V gets revised. This adjustment occurs based on information aggregated from the neighborhood $N(u)$ of node $u$ in the graph (as illustrated in Figure 5). This iterative process of message-passing forms a crucial part of GNN operations.

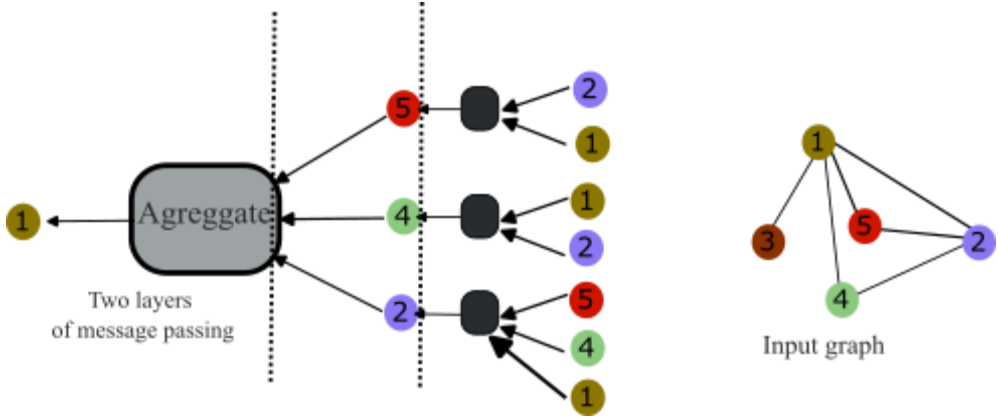

**Figure 5:** Message passing algorithm applied for a node.

In Figure 5, we see how a single node in a graph aggregates messages from its nearby neighbors. For instance, node 1 collects messages from its immediate neighbors (nodes 5, 4, and 2). What makes this process interesting is that these neighboring nodes, in turn, aggregate messages from their respective neighbors, creating a recursive pattern of information exchange. This visualization showcases a two-layer version of the message-passing model, illustrating how information ripples through the graph. Additionally, the computation graph forms a tree structure as it unfolds around the target node, depicting the hierarchical nature of message aggregation in Graph Neural Networks (GNNs).

Mathematically it could be expressed as follows:

$$h_u^{(k+1)} = Update^{(k)}(h_u^{(k)}, Aggregate^{(k)}(\{h_u^{(k)}, \forall \vartheta \in (u)\}))$$
$$= Update^{(k)}(h_u^{(k)}, m_{N(u)}^{(k)})$$

In the Graph Neural Network (GNN) framework, the processes of updating [14] node embeddings involve two main steps: AGGREGATE and UPDATE. Here, these are flexible, differentiable functions, often implemented as neural networks. During each iteration k of the GNN, the AGGREGATE function processes the embeddings of nodes in u's neighborhood $N(u)$, creating a message) $m_{N(u)}^{(k)}$ based on this collective neighborhood information. Subsequently, the UPDATE function combines this message $m_{N(u)}^{(k)}$ with the prior embedding $h_u^{(k-1)}$ of node u to generate the updated embedding $h_u^{(k)}$

Initially, at k=0, the embeddings are set to the input features for all nodes $h_u^{(0)}=x_u$, where xu represents the features of node u. After running K iterations of GNN message passing, the final layer's output $h_u^{(k)}$ represents the embeddings for each node in the graph $z_u=h_u^{(k)}$. This iterative process allows GNNs to capture complex relationships within graph-structured data.

## 4. Computational modules

In this part, we explore how node embeddings and message passing algorithms work together. We break down the mechanisms that these computer modules use to make graph-based learning powerful. They take raw data and turn it into important, detailed representations. By looking closely at these modules, we reveal the processes that make GNNs efficient and effective. This helps us understand better how GNNs can be useful in different areas.

Some commonly used computational modules are:

• **Propagation Module:** This module facilitates the flow of information between nodes, allowing the aggregated data to encompass both feature and topological details. Within propagation modules, the convolution and recurrent operators play key roles in gathering information from neighboring nodes, while the skip connection operation helps integrate insights from historical node representations, addressing concerns like over-smoothing.

• **Sampling Module:** For large graphs, the use of sampling modules becomes essential in the propagation process. These modules are typically integrated with the propagation module to ensure effective information dissemination across the graph.

• **Pooling Module:** When the focus shifts to obtaining representations of higher-level subgraphs or entire graphs, pooling modules come into play. These modules are instrumental in extracting valuable information from nodes to construct meaningful representations.

These computational modules form the building blocks of a typical Graph Neural Network (GNN) model, often combined to create the overall architecture. In the central portion of figure 6.a common GNN model design is depicted. Each layer incorporates the convolutional operator, recurrent operator, sampling module, and skip connection for effective information propagation [15]. Following this, a pooling module is introduced to extract high-level information. The stacking of these layers is a standard practice to enhance the quality of representations. While this architecture generally applies to most GNN models.

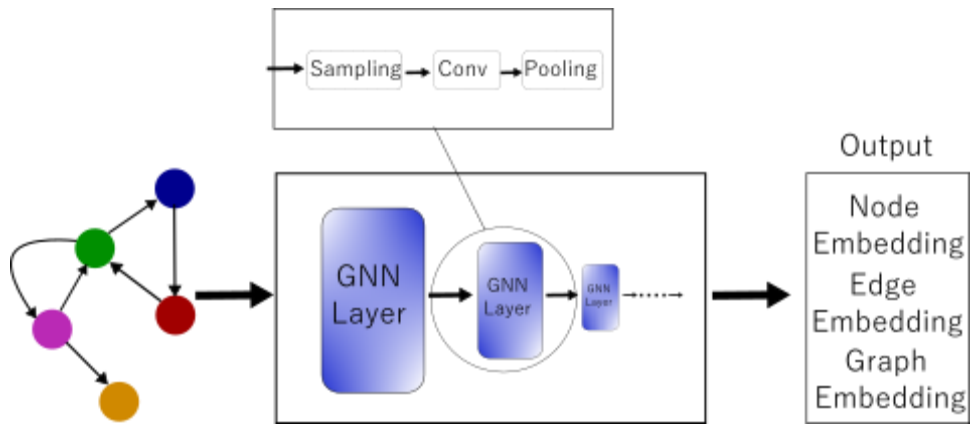

**Figure 6:** design pipeline for a GNN model

In a Graph Neural Network (GNN) layer, several key operations come into play. The sampling module is employed, particularly in large graphs, to facilitate the propagation of information. Simultaneously, the convolutional operator plays a pivotal role in gathering insights from neighboring nodes, capturing both features and topological information. Additionally, the introduction of the pooling module allows for the extraction of high-level information, crucial when dealing with representations of larger subgraphs or entire graphs. These operations within a GNN layer work together to improve the model's capacity to comprehend intricate graph structures.

## 5. Graph Generation

The objective of graph generation is to construct models capable of producing realistic graph structures. [16] Conceptually, this task can be linked to the inverse of the graph embedding problem. Rather than assuming a given graph structure G = (V, E) as input, in graph generation, our aim is for the model's output to be a graph G. While generating any arbitrary graph is relatively straightforward, such as a fully connected or edgeless graph, the true challenge lies in generating graphs with specific desired properties.

### 5.1. Traditional Graph Generation

Conventional methods for graph generation typically entail defining a generative process to describe the creation of edges within a graph. This generative process is often framed as establishing the probability or likelihood, denoted as P(A[u, v] = 1), of an edge existing between nodes u and v. The primary challenge lies in formulating a generative process that is both manageable and capable of producing graphs with non-trivial properties or characteristics. Manageability is crucial to facilitate sampling or analysis of the generated graphs. Simultaneously, we strive for these graphs to exhibit properties that align with real-world graph patterns.

### 5.2. Deep Generative Models

The conventional graph generation methods explored in the preceding chapter prove valuable across various scenarios. They excel in efficiently producing synthetic graphs with specified properties and offer insights into the potential emergence of certain graph structures in the real world. Nonetheless, a notable constraint of these traditional approaches lies in their dependence on fixed, manually designed generation processes. In essence, while these methods can generate graphs, they fall short in their capacity to autonomously learn a generative model from data [17].

We will introduce a series of basic deep generative models for graphs. These models will adapt three of the most popular approaches to building general deep generative models: variational autoencoders (VAEs) [18], generative adversarial networks (GANs) [19], as shown in figure 7 a standard VAE model applied to the graph setting. An encoder neural network maps the input graph $G = (A, X)$ to a posterior distribution $q\varphi(Z|X)$ over latent variables Z. Given a sample from this posterior, the decoder model $p\theta(X|Z)$ attempts to reconstruct the adjacency matrix.

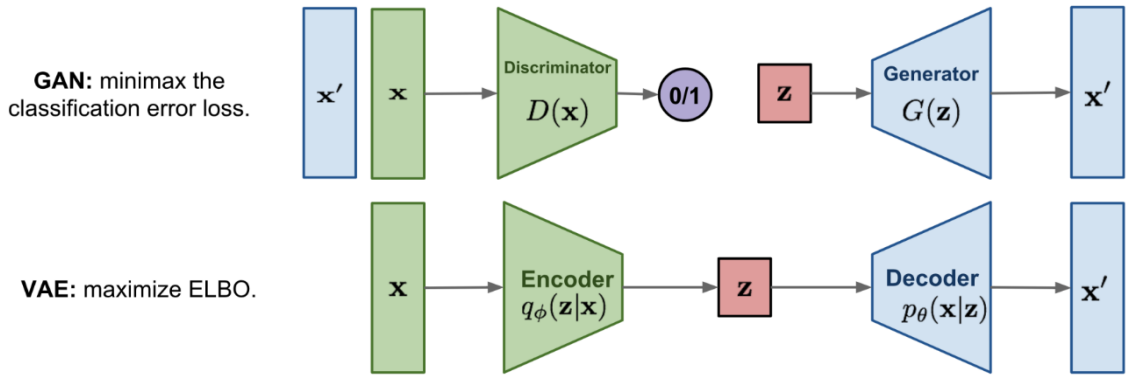

**Figure 7:** Comparison of two categories of generative models.

The initial objective is to adeptly interpret significant graphs from encoded latent representations, particularly when presented with training graphs. Meanwhile, the secondary objective serves as a regularization mechanism, guaranteeing our ability to interpret meaningful graphs even when latent representations are sampled from the prior distribution, denoted as $p(Z)$. This second goal assumes paramount importance, especially in scenarios where the generation of new graphs is desired post-training. The generation process relies on sampling from the prior and inputting these latent embeddings into the decoder, functioning effectively only when the second goal is met.

### 5.3. Evaluating Graph Generation

In the previous section, we explored more advanced graph generation methods based on Variational Autoencoders (VAEs) and Generative Adversarial Networks (GANs). While discussing these methods, we hinted at the effectiveness of some approaches compared to others and showcased examples of generated graphs in figure 8, illustrating the diverse capabilities of these approaches. However, determining which graph generation approach is superior poses a challenge. Unlike tasks with a clear notion of accuracy or error, evaluating generative models lacks a straightforward measure.

For instance, we could compare reconstruction losses or model likelihoods on held-out graphs, but this is complicated by the absence of a consistent likelihood definition across different generation approaches. In the realm of general graph generation, the current practice involves analyzing various statistics of the generated graphs and comparing their distribution to a test set.

Formally, let's assume we have a set of graph statistics denoted as $S = (s1, s2, \ldots, sn)$. Each of these statistics, represented as is $G : R \rightarrow [0, 1]$, is assumed to define a univariate distribution over the real numbers.

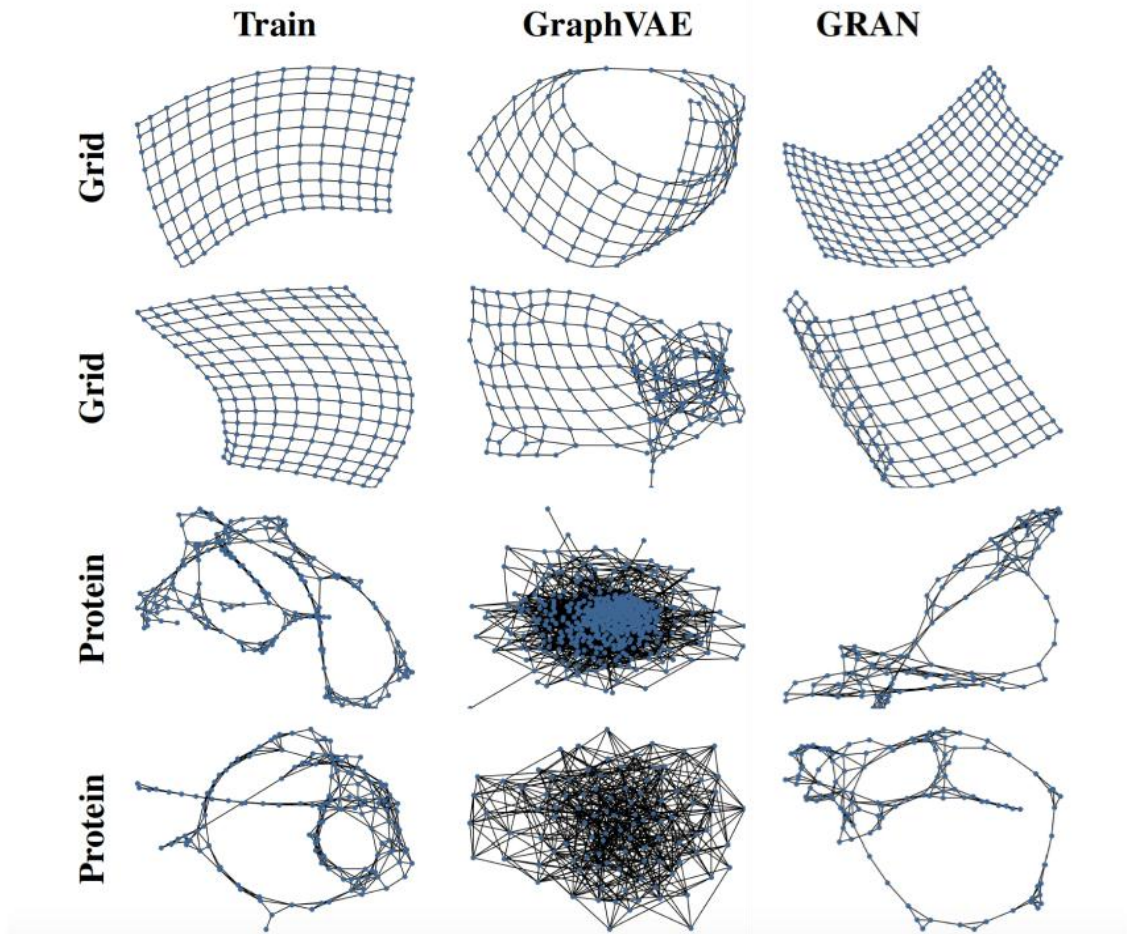

**Figure 8:** Examples of graphs generated by VAE and GRAN

Figure 8: Examples of graphs generated by a basic graph-level VAE as well as the GRAN model. Each row corresponds to a different dataset. The first column shows an example of a real graph from the dataset, while the other columns are randomly selected samples of graphs generated by the corresponding model [Liao et al., 2019a] [20]

### 5.4. Python Libraries for GNN's

There are many open-source libraries that allow the creation of deep neural network in Python, without having to explicitly write the code from scratch. In this section, we'll discover three of the
most popular: TensorFlow, Keras, and PyTorch. They all share some common features, as follows:
- The basic unit for data storage is the tensor. Consider the tensor as a generalization of a matrix to higher dimensions. Mathematically, the definition of a tensor is more complex, but in the context of deep learning libraries, they are multi-dimensional arrays of base values.
- Neural networks are represented as a computational graph of operations. The nodes of the graph represent the operations (weighted sum, activation function, and so on). The edges represent the flow of data, which is how the output of one operation serves as an input for the next one. The inputs and outputs of the operations (including the network inputs and outputs) are tensors.

*5.4.1    TensorFlow*

TensorFlow (TF) (https://www.tensorflow.org), is the most popular deep learning library. It's developed and maintained by Google. You don't need to explicitly require the use of a GPU, rather TensorFlow will automatically try to use it if you have one. If you have more than one GPU, you must assign operations to each GPU explicitly, or only the first one will be used. TensorFlow has a steeper learning curve, compared to the other libraries.

*5.4.2    Keras*

Keras is a high-level neural net Python library that runs on top of TensorFlow, CNTK (https://github.com/Microsoft/CNTK), or Theano., we'll assume that it uses TensorFlow on the backend. With Keras, you can perform rapid experimentation and it's relatively easy to use compared to TF. It will automatically detect an available GPU and attempt to use it. Otherwise, it will revert to the CPU.

*5.4.3    PyTorch:*

PyTorch (https://pytorch.org/) is a deep learning library based on Torch and developed by Facebook. It is relatively easy to use and has recently gained a lot of popularity. It will automatically select a GPU, if one is available, reverting to the CPU otherwise.

## 6.  Conclusion

In conclusion, this paper navigates through the intricate landscape of Graph Neural Networks (GNNs), shedding light on their applications, design principles, and emerging trends. It emphasizes the principal role of graphs as a universal language for understanding complex systems and illustrates how GNNs revolutionize graph analysis. The exploration encompasses essential concepts such as graph description, types, and scales, providing a solid groundwork. Dynamic operations and the temporal dimension of graphs are also explored, and the design pipeline of GNNs have been discussed, revealing the significance of computational modules in graph-based learning. The paper concludes by digging into the fascinating realm of graph generation, evaluating methods, and offering an overview of prominent Python libraries for GNNs. In essence, this paper serves as a comprehensive guide, illuminating the transformative capabilities of GNNs in deciphering intricate relationships within diverse graph-structured data.